\documentclass[runningheads]{llncs}
\usepackage[utf8]{inputenc}
\usepackage{amsmath}
\usepackage{amssymb}
\usepackage{mathtools}
\usepackage{xcolor}
\usepackage{svg}
\usepackage{calc}
\usepackage{hyperref}
\usepackage[ruled,vlined,linesnumbered,noend]{algorithm2e}
\usepackage[margin=0.3cm]{caption}
\pdfoutput=1
\title{Probabilistic Guarantees for \\ Safe Deep Reinforcement Learning}
\author{Edoardo Bacci \orcidID{0000-0002-0367-898X}
\and David Parker \orcidID{0000-0003-4137-8862}}
\institute{University of Birmingham, Birmingham, United Kingdom \\
\email{exb461@bham.ac.uk, d.a.parker@cs.bham.ac.uk}}

\renewcommand{\emptyset}{\varnothing}

\usepackage{scalerel}

\spnewtheorem{assumption}{Assumption}{\bfseries}{\itshape}
\newcounter{exampcount}
\newcommand{\startpara}[1]{{%
\vskip6pt\noindent
{\bf #1.}}}

\def\matr#1{{\mathbf{#1}}}

\def\Nset{\mathbb{N}}

\def\Rset{\mathbb{R}}

\renewcommand{\leq}{\leqslant}
\renewcommand{\geq}{\geqslant}
\renewcommand{\le}{\leqslant}
\renewcommand{\ge}{\geqslant}

\def\squareforqed{\hbox{\rlap{$\sqcap$}$\sqcup$}}
\def\qed{\ifmmode\squareforqed\else{\unskip\nobreak\hfil
\penalty50\hskip1em\null\nobreak\hfil\squareforqed
\parfillskip=0pt\finalhyphendemerits=0\endgraf}\fi}

\newcommand\dist{{\mathit{Dist}}}
\newcommand\Dist{{\dist}}
\newcommand{\supp}{\mathit{Supp}}

\newcommand\Path{{\mathit{Path}}}

\renewcommand{\Pr}[2]{{\mathit{Pr}_{#1}^{#2}}}

\def\AP{{\mathit{AP}}}

\def\Act{{A}}
\def\act{a}

\newcommand{\fail}{{\mathit{fail}}}

\newcommand{\dave}[1]{\marginpar{\footnotesize \color{teal} {\bf D:} \textsf{#1}}}

\SetAlgoSkip{}

\def\techreport{}

\begin{document}

\maketitle
\begin{abstract}
Deep reinforcement learning has been successfully applied to many control tasks,
but the application of such controllers in safety-critical scenarios has been limited
due to safety concerns. 
Rigorous testing of these controllers is challenging,
particularly when they operate in probabilistic environments
due to, for example, hardware faults or noisy sensors.
We propose MOSAIC, an algorithm for measuring the safety of
deep reinforcement learning controllers in stochastic settings.
Our approach is based on the iterative construction of a formal abstraction
of a controller's execution in an environment,
and leverages probabilistic model checking of Markov decision processes to
produce probabilistic guarantees on safe behaviour over a finite time horizon.
It produces bounds on the probability of safe operation of the controller
for different initial configurations and identifies regions where
correct behaviour can be guaranteed.
We implement and evaluate our approach on controllers trained for
several benchmark control problems.
%

\end{abstract}


\section{Introduction}

Deep reinforcement learning is the application of deep neural networks
to solve reinforcement learning tasks.
This technique has been shown to solve many complex control tasks successfully \cite{Bougiouklis2018,Shalev-Shwartz2016,Ohn-Bar2016}.
However, real-world applications of these methods,
especially in safety-critical scenarios such as autonomous driving,
is limited because it is difficult to establish guarantees on their safety.

Formal verification is a rigorous approach to checking the correctness of computerised systems.
It is particularly appealing for systems that are based on neural networks,
because the training process often yields models that are large, complex and opaque.
Furthermore, the input space is typically too large to allow exhaustive testing,
and there now exist a variety of approaches to construct adversarial attacks,
i.e., small and imperceptible perturbations to the inputs of the neural network
that cause it to produce erroneous outputs.

In recent years, there has been growing interest in verification techniques
for neural networks \cite{Katz2017,HKW+17,TimonGehr2018},
with a particular focus on the domain of image classification.
These aim to prove the absence of particular classes of adversarial attack,
typically those that are ``close'' to inputs for which the correct output is known.
Methods proposed include mapping the verification to an SMT (satisfiability modulo theories) problem
and the use of abstract interpretation.

There are also various approaches to tackling safety in reinforcement learning.
For example, safe reinforcement learning \cite{GF15} factors in safety objectives
into the learning process.
Using formal specifications of the objectives has also been proposed,
such as maximising the probability of satisfying a temporal logic objective~\cite{BCC+14,FuT14,DBLP:conf/tacas/HahnPSSTW19}
or restricting learning to a set of verified policies~\cite{JJD+16}.
More recently, formal verification of deep reinforcement learning systems
has been considered~\cite{KazakBKS19}, by leveraging existing neural network verification methods.

A further challenge for verifying the safe operation of controllers
synthesised using deep reinforcement learning is the fact they
are often developed to function in uncertain or unpredictable environments.
This necessitates the use of stochastic models
to train, and to reason about, the controllers.
One source of probabilistic behaviour is dynamically changing environments
and/or unreliable or noisy sensing.
Another source, and the one we focus on here, is the occurrence of faults,
e.g., in the hardware for actuators in the controller.

In this paper, we propose novel techniques to establish \emph{probabilistic} guarantees
on the safe behaviour of deep reinforcement learning systems
which can be subject to faulty behaviour at runtime.
Our approach, which we call MOSAIC (MOdel SAfe Intelligent Control)
uses a combination of abstract interpretation and probabilistic verification
to synthesise the guarantees.

Formally, we model the runtime execution of a deep reinforcement learning based controller
as a continuous-space discrete-time Markov processes (DTMP).
This is built from: (i) the neural network specifying the controller;
(ii) a controller fault model characterising the probability with which faults
occur when attempting to execute particular control actions;
and (iii) a deterministic, continuous-space model of the physical environment,
which we assume to be known.

We concern ourselves with finite-horizon safety specifications
and consider the probability with which a failure state is reached
within a specified number of time steps.
More precisely, our main aim is to identify ``safe'' regions of the possible initial
configurations of the controller, for which this failure probability
is guaranteed to be below some specified threshold.

One key challenge to overcome, due to the continuous-space model,
is that the number of initial configurations is infinite.
We construct a finite-state abstraction 
as a Markov decision process (MDP), comprising abstract states (based on intervals) that represent
regions of the state space of the concrete controller model.
We then use standard probabilistic model checking techniques on the MDP abstraction,
and show that this yields upper bounds on the step-bounded failure probabilities
for different initial regions of the controller model.

A second challenge is that constructing the abstraction
requires extraction of the controller policy from its neural network representation.
We perform a symbolic analysis of the neural network,
for which we design a branch-and-bound algorithm,
and an abstraction process that explores the reachable
abstract states of the environment.
We also iteratively refine the abstraction to
yield more accurate bounds on the failure probabilities.
We evaluate our approach by applying it to deep reinforcement learning controllers
for two benchmark control problems: a cartpole and a pendulum.




\startpara{Related Work}
As discussed above, various verification techniques for neural networks exist,
including those based on abstract interpretation.
Some use abstractions based on intervals~\cite{Ruan2018,Wang2018,Anderson2019},
as we do; others use more sophisticated representations
such as polyhedra and zonotopes~\cite{TimonGehr2018}.
Recently, correctness for Bayesian neural networks has been considered, using probabilistic notions of robustness, e.g.,~\cite{cardelli2019statistical}.
Mostly, these approaches focus on supervised learning, often for image classification,
but they have also been built upon for verified deep reinforcement learning~\cite{KazakBKS19},
where (non-probabilistic) safety and liveness properties are checked.
Other, non-neural network based, reinforcement learning has also been verified,
e.g., by extracting and analysing decision trees \cite{bastani18}.

In the context of probabilistic verification,
neural networks have been used to find POMDP policies
with guarantees~\cite{CJW+19,CJT20},
but with recurrent neural networks and for discrete, not continuous, state models.
Also related are techniques to verify continuous space probabilistic
models, e.g., \cite{LAB15,SGA15} which build finite-state abstractions
as Markov chains or interval Markov chains.
Finally, there is a large body of work on abstraction for probabilistic verification;
ours is perhaps closest in spirit to the game-based abstraction approach for MDPs from \cite{KKNP10}.

\section{Preliminaries}

We will use $\Dist(X)$ to denote the set of discrete probability distributions over the set $X$,
i.e., functions $\mu:X\to [0,1]$ where $\sum_{x\in X}\mu(x)=1$.
The support of $\mu$, denoted $\supp(\mu)$,
is defined as $\supp(\mu)=\{x\in X\,|\,\mu(x)>0\}$.
In some cases, we will use distributions where the set $X$ is uncountable but where the support is finite.
We also write $\mathcal{P}(X)$ to denote the powerset of $X$.

We use two probabilistic models: 
\emph{discrete-time Markov processes} (DTMPs) to model controllers, 
and \emph{Markov decision processes} (MDPs) for abstractions.


\begin{definition}[Discrete-time Markov process]\label{def:dtmp}
A (finite-branching) \emph{discrete-time Markov process}
is a tuple $(S,S_0,\matr{P},\AP,L)$, where:
$S$ is a (possibly uncountably infinite) set of states;
$S_0\subseteq S$ is a set of initial states;
$\matr{P}:S\times S\to [0,1]$ is a transition probability matrix,
where $\sum_{s'\in \supp(\matr{P}(s,\cdot))}\matr{P}(s,s')=1$ for all $s\in S$;
$\AP$ is a set of atomic propositions; and $L:S\to\AP$ is a labelling function.
\end{definition}

The process starts in some initial state $s_0\in S_0$
and then evolves from state to state in discrete time steps.
When in state $s$, the probability of making a transition to state $s'$ is given by $\matr{P}(s,s')$.
We assume that the process is finite-branching, i.e., the number of possible successors of each state
is finite, despite the continuous state space.
This simplifies the representation and suffices for the probabilistic behaviour that we model in this paper.

A \emph{path} is an infinite sequence of states $s_0 s_1 s_2\dots$ through the model,
i.e., such that 
$\matr{P}(s_i,s_{i+1})>0$ for all $i$.
We write $\Path(s)$ for the set of all paths starting in a state $s$.
In standard fashion~\cite{KSK76}, we can define a probability space $\Pr{s}{}$ over $\Path(s)$.
Atomic propositions from the set $\AP$ will be used to specify properties for verification;
we write $s\models b$ for $b\in\AP$ if $b\in L(s)$.


\begin{definition}[Markov decision process]
A \emph{Markov decision process} is a tuple $(S,S_0,\matr{P},\AP,L)$, where:
$S$ is a finite set of states;
$S_0\subseteq S$ are initial states;
$\matr{P}:S\times \Nset\times S\to [0,1]$ is a transition probability function,
where $\sum_{s'\in S}\matr{P}(s,j,s')\in\{0,1\}$ for all $s\in S,j\in\Nset$;
$\AP$ is a set of atomic propositions; and $L:S\to\AP$ is a labelling function.
\end{definition}

Unlike discrete-time Markov processes above,
we assume a finite state space.
A transition in a state $s$ of an MDP first requires a choice between
(finitely-many) possible probabilistic outcomes in that state.
Unusually, we do not use action labels to distinguish these choices,
but just integer indices. Primarily, this is to avoid confusion with the use of actions
taken by controllers, which do not correspond directly to these choices.
The probability of moving to successor state $s'$ when taking choice $j$ in state $s$
is given by $\matr{P}(s,j,s')$.

As above, a path is an execution through the model,
i.e., an infinite sequence of states and indices $s_0 j_0 s_1 j_1\dots$
such that $\matr{P}(s_i,j_i,s_{i+1})>0$ for all $i$.
%
A \emph{policy} of the MDP selects the choice to take in each state, based on the history
of its execution so far.
For a policy $\sigma$, we have a probability space $\Pr{s}{\sigma}$
over the set of paths starting in state $s$.
If $\psi$ is an event of interest defined by a measurable set of paths
(e.g., those reaching a set of target states),
we are usually interested in the minimum or maximum probability
of the event over all policies:

$$
\Pr{s}{\min}(\psi) = \inf_\sigma \Pr{s}{\sigma}(\psi)
\ \ \ \mbox{ and } \ \ \ 
\Pr{s}{\max}(\psi) = \sup_\sigma \Pr{s}{\sigma}(\psi)
$$

\section{Controller Modelling and Abstraction}\label{sec:maths}

In this section, we formalise the problem of modelling and verifying deep reinforcement learning controllers,
and then describe the MDP abstraction
that underlies our MOSAIC approach to performing the verification.


\subsection{Controller Execution Model}

We consider controllers acting over continuous state spaces systems with a discrete action space.
We assume a set of $n$ real-valued state space variables 
and denote the state space by $S = \Rset^n$.
There is a finite set $\Act=\{\act_1,\dots,\act_m\}$
of $m$ actions that can be taken by the controller.
For simplicity, we assume that all actions are available in every state.

To describe the execution of a controller, we require three things:
(i) a \emph{controller policy}; (ii) an \emph{environment model};
and (iii) a \emph{controller fault model}.
Each is described in more detail below.

\begin{definition}[Controller policy]
A \emph{controller policy} is a function $\pi : S\to \Act$,
which selects an action $\pi(s)$ for the controller to take in each state $s\in S$.
\end{definition}

We restrict our attention to policies that are
memoryless (choosing the same action in each state $s$) and
deterministic (selecting a fixed single action, with no randomisation).
In this work, policies are represented by neural networks,
and generated through deep reinforcement learning.
However, for the purposes of this section,
we treat the policy simply as a function from states to actions.

\begin{definition}[Environment model]
An \emph{environment model} is a function $E: S\times \Act \to S$
that describes the state $E(s,\act)$ of the system after one
time step if controller action $\act$ is (successfully) taken in state $s$.
\end{definition}

The environment represents the effect that each action executed by a controller has on the system.
We assume a deterministic model of the environment;
probabilistic behaviour due to failures is introduced separately (see below).

We also extend $E$ to define the change in system state when a \emph{sequence} of
zero or more actions are executed, still within a single time step.
This will be used below to describe the outcome of controller execution faults.
Re-using the same notation, for state $s\in S$ and action sequence $w\in\Act^*$,
we write $E(s,w)$ to denote the outcome of taking actions $w$ in $s$.
This can be defined recursively:
for the empty action sequence $\epsilon$,
we have $E(s,\epsilon)=s$; and, for a sequence of $k$ actions $\act_1\dots\act_k$, we have
$E(s,\act_1\dots\act_k) = E(E(s,\act_1\dots\act_{k-1}),\act_k)$.

\begin{definition}[Controller fault model]
A \emph{controller fault model} is a function $f:\Act\to\Dist(\Act^*)$
that gives, for each possible controller action,
the sequences of actions that may actually result and their probabilities.
\end{definition}

This lets us model a range of controller faults.
A simple example is the case of an action $\act$ failing to execute with some
probability $p$: we have $f(\act)(\epsilon)=p$, $f(\act)(\act)=1{-}p$ and
$f(\act)(w)=0$ for all other action sequences $w$.
Another example, is a ``sticky'' action \cite{machado2018revisiting} $\act$ which executes twice with probability $p$,
i.e., $f(\act)(\act\act)=p$, $f(\act)(\act)=1{-}p$ and $f(\act)(w)=0$ for any other $w$.

Now, given a controller policy $\pi$, an environment model $E$
and a controller fault model $f$, we can formally define the behaviour
of the execution of the controller within the environment.
We add two further ingredients:
a set $S_0\subseteq S$ of possible \emph{initial states};
and a set $S_\fail\subseteq S$ of \emph{failure states},
i.e., states of the system where we consider it to have failed.
We refer to the tuple $(\pi,E,f,S_0,S_\fail)$ as a \emph{controller execution}.
Its \emph{controller execution model}
is a (continuous-space, finite-branching) discrete-time Markov process defined as follows.

\begin{definition}[Controller execution model]\label{def:ctrlmodel}
Given a controller execution $(\pi,E,f,S_0,S_\fail)$, the corresponding
\emph{controller execution model} describing its runtime behaviour
is the DTMP $(S,S_0,\matr{P},\AP,L)$ where
$AP=\{\fail\}$, for any $s\in S$, $\fail\in L(s)$ iff $s\in S_\fail$ and,
for states $s,s'\in S$:
$$
\matr{P}(s,s') = \sum\left\{ f(\pi(s))(w) \ | \ w\in\Act^* \mbox{ s.t. } E(s,w)=s' \right\}.
$$
\end{definition}
For each state $s$, the action chosen by the controller policy is $\pi(s)$
and the action sequences that may result are given by
the support of the controller fault model distribution $f(\pi(s))$.
For each action sequence $w$, the resulting state is $E(s,w)$.
In the above, to define $\matr{P}(s,s')$
we have combined the probability of all such sequences $w$ that lead to $s'$
since there may be more than one that does so.


Recall the example controller fault models described above.
For an action $\act$ that fails to be executed with probability $p$,
the above yields $\matr{P}(s,s)=p$ and $\matr{P}(s,E(s,\act))=1{-}p$.
For a ``sticky'' action $\act$ (with probability $p$ of sticking), it yields
$\matr{P}(s,E(E(s,\act),\act))=p$ and $\matr{P}(s,E(s,\act))=1{-}p$.



\subsection{Controller Verification}\label{sec:verif}

Using the model defined above of a controller operating in a given environment,
our aim is to verify that it executes safely.
More precisely, we are interested in the probability of
reaching \emph{failure states} within a particular time horizon.
We write $\Pr{s}{}(\Diamond^{\leq k}\mathit{fail})$ for the probability of reaching a failure state
within $k$ time steps when starting in state $s$, which can be defined as:
$$
\Pr{s}{}(\Diamond^{\leq k}\mathit{fail})
= \Pr{s}{}(\{s_0 s_1 s_2\dots \in\Path(s)\,|\,s_i\models\mathit{fail}\mbox{ for some } 0\leq i \leq k\})
$$
Since we work with discrete-time, finite-branching models,
we can compute finite-horizon reachability probabilities recursively as follows:
$$
\Pr{s}{}(\Diamond^{\leq k}\mathit{fail})=
\left\{\begin{array}{cl}
1 & \mbox{if } s \models\mathit{fail} \\
0 & \mbox{if } s \not\models\mathit{fail} \land k{=}0 \\
\sum_{s'\in \supp(\matr{P}(s,\cdot))}\matr{P}(s,s')\cdot\Pr{s'}{}(\Diamond^{\leq k-1}\mathit{fail}) \ \ & \mbox{otherwise.}
\end{array}\right.
$$
For our controller execution models, we are interested in two closely related verification problems.
First, for a specified probability threshold $p_\mathit{safe}$, we would like
to determine the subset $S_0^\mathit{safe}\subseteq S_0$ of ``safe'' initial states
from which the error probability is below the threshold:
$$
S_0^\mathit{safe} = \{s\in S_0 \ | \ \Pr{s}{}(\Diamond^{\leq k}\mathit{fail}) < p_\mathit{safe}\}
$$
Alternatively, for some set of states $S'$, typically the initial state set $S_0$,
or some subset of it, we wish to know the maximum (worst-case) error probability:
$$
p_{S'}^{+} = \sup \{\Pr{s}{}(\Diamond^{\leq k}\mathit{fail}) \ | \ s\in S'\}
$$
This can be seen as a \emph{probabilistic guarantee} over the executions
that start in those states.
In this paper, we tackle approximate versions of these problems,
namely under-approximating $S_0^\mathit{safe}$
or over-approximating $p_{S'}^{+}$.



\subsection{Controller Execution Abstraction}\label{sec:abstraction}

A key challenge in tackling the controller verification problem outlined above
is the fact that it is over a continuous-state model.
In fact, since the model is finite-branching and we target finite-horizon safety properties,
for a specific initial state, the $k$-step probability of a failure could
be computed by solving a finite-state Markov chain.
However, we verify the controller for a \emph{set} of initial states,
giving infinitely many possible probabilistic executions.

Our approach is to construct and solve an \emph{abstraction} of the model of controller execution.
The abstraction is a finite-state MDP whose states are 
\emph{abstract states} $\hat{s} \subseteq S$, each representing
some subset of the states of the original concrete model.
We denote the set of all possible abstract states as $\hat{S}\subseteq\mathcal{P}(S)$.
In our approach, we use intervals (i.e., the ``Box'' domain; see Section~\ref{sec:absdom}).

In order to construct the abstraction of the controller's execution,
we build on an abstraction $\hat{E}$ of the environment 
$E:S\times\Act\to S$.
This abstraction is a function $\hat{E}:\hat{S}\times\Act\to\hat{S}$
which soundly over-approximates the (concrete) environment,
i.e., it satisfies the following definition.
\begin{definition}[Environment abstraction]\label{def:envabs}
For environment model $E:S\times\Act\to S$ and set of abstract states $\hat{S}\subseteq\mathcal{P}(S)$,
an \emph{environment abstraction} is a function $\hat{E}:\hat{S}\times\Act\to\hat{S}$ such that:
for any abstract state $\hat{s}\in\hat{S}$, concrete state $s\in\hat{s}$ and action $\act\in\Act$,
we have $E(s,\act) \in \hat{E}(\hat{s},\act)$.
\end{definition}
Using interval arithmetic, we can construct $\hat{E}$ for a wide range of functions $E$.
As for $E$, the environment abstraction $\hat{E}$ extends naturally to action sequences,
where $\hat{E}(\hat{s},w)$ gives the result of taking a sequence $w$ of actions in abstract state $\hat{s}$.
It follows from Definition~\ref{def:envabs} that,
for any abstract state $\hat{s}\in\hat{S}$, concrete state $s\in\hat{s}$ and action sequence $w\in\Act^*$,
we have $E(s,w) \in \hat{E}(\hat{s},w)$.

Our abstraction is an MDP whose states are abstract states from the set $\hat{S}\subseteq\mathcal{P}(S)$.
This represents an over-approximation of the possible behaviour of the controller,
and computing the maximum probabilities of reaching failure states in the MDP
will give upper bounds on the actual probabilities in the concrete model.
The choices that are available in each abstract state $\hat{s}$ of the MDP
are based on a partition of $\hat{s}$ into subsets $\{\hat{s}_1,\dots,\hat{s}_m\}$.
Intuitively, each choice represents the behaviour for states in the different subsets $\hat{s}_j$.


\begin{definition}[Controller execution abstraction]\label{def:ctrlabs}
For a controller execution $(\pi,E,f,S_0,S_\fail)$,
a set $\hat{S}\subseteq\mathcal{P}(S)$ of abstract states
and a corresponding environment abstraction $\hat{E}$,
the \emph{controller execution abstraction}
is defined as an MDP $(\hat{S},\hat{S}_0,\hat{\matr{P}},\AP,\hat{L})$ satisfying the following:
\begin{itemize}
\item for all $s\in S_0$, $s\in\hat{s}$ for some $\hat{s}\in \hat{S}_0$;
\item for each $\hat{s}\in\hat{S}$, there is a partition $\{\hat{s}_1,\dots,\hat{s}_m\}$ of $\hat{s}$
that is \emph{consistent} with the controller policy $\pi$ (i.e.,
$\pi(s)=\pi(s')$ for any $s,s'\in \hat{s}_j$ for each $j$)
and, for each $j\in\{1,\dots,m\}$ we have:
$$
\hat{\matr{P}}(\hat{s},j,\hat{s}') = \sum\left\{ f(\pi(\hat{s}_j))(w) \ | \ w\in\Act^* \mbox{ such that } \hat{E}(\hat{s}_j,w)=\hat{s}' \right\}
$$
where $\pi(\hat{s}_j)$ is the action that $\pi$ chooses for all states $s\in \hat{s}_j$;
\item $\AP=\{\fail\}$ and $\fail\in \hat{L}(\hat{s})$ iff $\fail\in L(s)$ for some $s\in\hat{s}$.
\end{itemize}
\end{definition}
The idea is that each $\hat{s}_j$ within abstract state $\hat{s}$
represents a set of concrete states that have the same behaviour at this level of abstraction.
This is modelled by the $j$th choice from $\hat{s}$,
which we construct by finding the controller action $\pi(\hat{s}_j)$ taken in those states,
the possible action sequences $w$ that may arise when taking $\pi(\hat{s}_j)$
due to the controller fault model $f$,
and the abstract states $\hat{s}'$ that result when applying $w$ in $\hat{s}_j$
according to the abstract model $\smash{\hat{E}}$ of the environment.

The above describes the general structure of the abstraction;
in practice, it suffices to construct a fragment of at most depth $k$ from the initial states.
Once constructed, computing maximum probabilities for the MDP yields upper bounds
on the probability of the controller exhibiting a failure.
In particular, we have the following result (see
\ifthenelse{\isundefined{\techreport}}{%
\cite{BP20arxiv}
}{%
the appendix
}%
for a proof):

\begin{theorem}\label{thm:abstr}
Given a state $s\in S$ of a controller model DTMP,
and an abstract state $\hat{s}\in \hat{S}$ of the corresponding controller abstraction MDP
for which $s\in\hat{s}$, we have
$\Pr{s}{}(\Diamond^{\leq k}\mathit{fail}) \ \leq \ \Pr{\hat{s}}{\max}(\Diamond^{\leq k}\mathit{fail})$.
\end{theorem}
This also provides a way to determine sound approximations for
the two verification problems discussed in Section~\ref{sec:verif},
namely finding the set $S_0^\mathit{safe}$
of states considered ``safe'' for a particular probability threshold $p_\mathit{safe}$:
$$
S_0^\mathit{safe} \ \supseteq \ 
\{s\in \hat{s} \ | \ \hat{s}\in\hat{S}_0 \mbox{ and } \Pr{\hat{s}}{\max}(\Diamond^{\leq k}\mathit{fail}) < p_\mathit{safe}\}
$$
and the worst-case probability $p_{S'}^{+}$ for a set of states $S'$:
$$
p_{S'}^{+} \ \leq \ 
\max \{\Pr{\hat{s}}{\max}(\Diamond^{\leq k}\mathit{fail}) \ | \ \hat{s}\in\hat{S} \mbox{ such that } \hat{s}\cap S'\neq\emptyset\}
$$


\section{Policy Extraction and Abstraction Generation}\label{sec:algs}

Building upon the ideas in the previous section,
we now describe the key parts of the MOSAIC algorithm to implement this.
We explain the abstract domain used, how to extract a controller policy
over abstract states from a neural network representation,
and then how to build this into a controller abstraction.
We also discuss data structures for efficient manipulation of abstract states.


\startpara{Abstract domain}\label{sec:absdom}
The abstraction described in Section~\ref{sec:abstraction} assumes an arbitrary
set of abstract states $\hat{S}\subseteq\mathcal{P}(S)$.
In practice, our approach assumes $S\subseteq\Rset^n$ and uses the ``Box'' abstract domain,
where abstract states are conjunctions of intervals (or hyperrectangles),
i.e., abstract states are of the form $[l_1,u_1]\times\cdots\times[l_n,u_n]$,
where $l_j,u_i\in\Rset$ are lower and upper bounds for $1\leq i\leq n$.

\subsection{Neural Network Policy Extraction}

Controller policies are functions $\pi:S\to\Act$, represented
as neural networks. To construct an abstraction (see Definition~\ref{def:ctrlabs}),
we need to divide abstract states into subregions which are \emph{consistent} with $\pi$,
i.e., those where $\pi(s)$ is the same for each state $s$ in the subregion.
Our overall approach is as follows. For each action $\act$, we first modify the neural network,
adding an \emph{action layer} to help indicate the states (network inputs) where $\act$ is chosen.
Then, we adapt a branch-and-bound style optimisation algorithm to identify these states,
which builds upon methods to approximate neural network outputs by propagating intervals through it.

\startpara{Branch and bound}
Branch and bound (BaB) is an optimisation algorithm which aims to minimise (or maximise) a given objective function. It works iteratively, starting from the full domain of possible inputs.
BaB estimates a maximum and minimum value for the domain using estimator functions,
which are quick to compute and approximate the real objective function by providing an \textit{upper bound} (UB) and a \textit{lower bound} (LB) between which the real function lies.
The chosen bounding functions must be admissible,
meaning we can guarantee that the real function will always lie within those boundaries.

At each iteration of BaB, the domain is split (or ``branched") into multiple parts.
In the absence of any additional assumptions about the objective function,
the domain is split halfway across the largest dimension.
For each part, the upper and lower bounds are calculated and regions whose lower bounds are higher than the current global minimum upper bound (the minimum amongst all regions' upper bounds) are discarded because, thanks to the admissibility property of the approximate functions, they cannot ever have a value lower than the global minimum upper bound.

The algorithm proceeds by alternating the branching phase and the bounding phase until the two boundaries converge or the difference between the bounds is less than an acceptable error value. After that, the current region is returned as a solution to the optimisation problem, and the algorithm terminates.

\begin{figure}[!t]
    \vspace*{-0.3cm}
	\centering
	\includegraphics[width=1\linewidth]{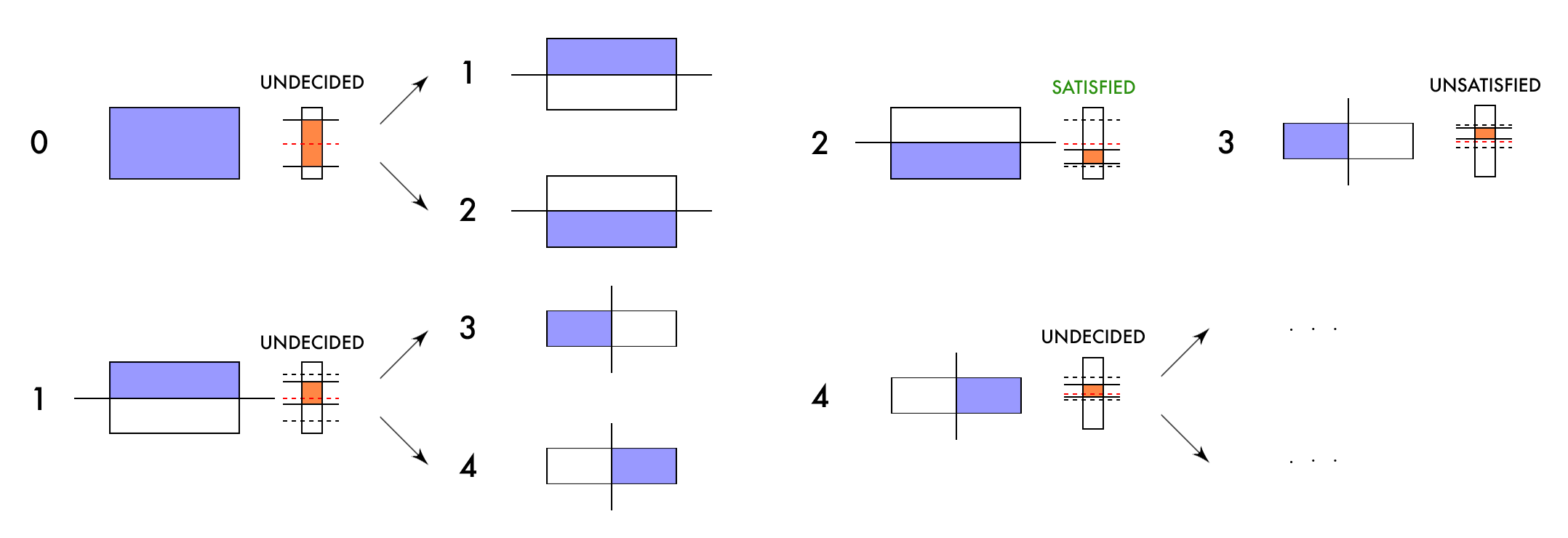}
    \vspace*{-0.4cm}
	\caption{
	    Illustrating branch-and-bound to identify actions. Each box represents an abstract state and the bar on the right represents upper and lower bounds on the output of the network.
		0) The upper and lower bounds of the domain do not give a definite answer, the domain is split into two subregions;
		1) The boundaries are tighter than in the previous iteration but the subregion is still undecided;
		2) The upper bound is $<$ 0, the property ``action taken is $\act$" is always true in this subregion;
		3) The lower bound is $>$ 0, the property ``action taken is $\act$" is always false in this subregion;
		4) The interval between upper and lower bound still contains 0, the action taken in this interval is still unknown so we continue to branch.
	}
	\label{fig:branch}
\end{figure}

\startpara{Finding consistent regions}
In order to frame the problem of identifying areas of the domain
that choose an action $\act$ as an optimisation problem,
we construct an additional layer that we call an ``action layer",
and append it on top of the neural network architecture.
This is built in such a way that the output is strictly negative if the 
output is $\act$, and strictly positive value if not.
We adopt the construction from \cite{Bunel2017}, which uses a layer to encode a correctness
property to be verified on the output of the network.

The techniques of~\cite{Bunel2017} also adapt branch-and-bound algorithms,
using optimisation to check if a correctness property is true.
But our goal is different: identifying areas within abstract states where action $\act$
is chosen, so we need a different approach.
Rather than minimising the modified output of the neural network,
we continue splitting domains
until we find areas that consistently either do or do not choose action $\act$ or we reach a given precision.
We do not keep track of the global upper or lower bound since we only need to consider
the local ones to determine which actions are taken in each subregion.
In the modified branch-and-bound algorithm, after calculating upper and lower bounds
for an interval, we have 3 cases:
\begin{itemize}
    \item $\mathit{UB} > \mathit{LB} > 0$ : the controller will never choose action $\act$ for the interval;
    \item $0 > \mathit{UB} > \mathit{LB}$ : the controller will always choose action $\act$; 
    \item $\mathit{UB} > 0 > \mathit{LB}$ : the outcome of the network is still undecided, so we split the interval and repeat for each sub-interval.
\end{itemize}
At the end of the computation, we will have a list of intervals which satisfy the property ``the controller always take action $\act$" and intervals which always violate it. From these two lists we can summarise the behaviour of the controller within the current region of the state space.

\begin{algorithm}[t]
    \DontPrintSemicolon
    \caption{Finding subregions of abstract state $\hat{s}$ for action $\act$}
    \label{alg:bab}
    \SetAlgoLined
    \SetKwFunction{FMain}{find\_action\_subregions}
    \SetKwFunction{computeUB}{compute\_UB}
    \SetKwFunction{computeLB}{compute\_LB}
    \SetKwFunction{modifynet}{add\_action\_layer}
    \SetKwFunction{split}{split}
    \SetKwProg{Fn}{function}{:}{}
    \Fn{\FMain{$net$, $\act$, $\hat{s}$}}{
        $queue$ $=$ \{$\hat{s}$\}, $sat$ $=$ \{ \}, $unsat$ $=$ \{ \}\;
        $mod\_net$ = \modifynet($net$, $\act$)\;
        \While{queue $\neq$ $\varnothing$}{
            $curr\_domain$ = $queue$.pop()\;
            $\mathit{UB}$ $=$ \computeUB($mod\_net$, $curr\_domain$)\;
            $\mathit{LB}$ $=$ \computeLB($mod\_net$, $curr\_domain$)\;
            \uIf{$\mathit{UB}< 0$}{
                $sat$.append($curr\_domain$)\;
            }
            \uElseIf{$\mathit{LB}>0$}{
                 $unsat$.append($curr\_domain$)\;
            }
            \Else{
                $dom_1$, $dom_2$ = \split($curr\_domain$)\;
                $queue$.append($dom_1$)\;
                $queue$.append($dom_2$)\;
            }
        }
        \KwRet sat,unsat\;
    }
\end{algorithm}

Algorithm~\ref{alg:bab} shows pseudocode for the overall procedure
of splitting an abstract state $\hat{s}$ into a set of subregions where an action $\act$
is always taken, and a set where it is not.
Figure~\ref{fig:branch} illustrates the algorithm executing for
a 2-dimensional input domain.
The blue subregions are the ones currently being considered;
the orange bar indicates the range between computed lower and upper bounds
for the output of the network, and the red dashed line denotes the zero line.


\startpara{Approximating neural network output}
The branch-and-bound algorithm requires computation of upper and lower bounds on the
neural network's output for a specific domain
($\mathtt{compute\_UB}$ and $\mathtt{compute\_LB}$ in Algorithm~\ref{alg:bab}).
To approximate the output of the neural network, we use the \textit{Planet} approach from~\cite{Ehlers2017}. The problem of approximating the output of the neural network lies in determining the output of the non-linear layers, which in this case are composed of ReLU units. ReLU units can be seen as having 2 phases: one where the output is a constant value if the input is less than 0 and the other where the unit acts as the identity function.
The algorithm tries to infer the phase of the ReLU function (whether \(x<0\) or \(x\ge0\)) by constraining the range of values from the input of the previous layers. In the case of the algorithm not being able to determine the phase of the activation function, some linear over-approximation boundaries are used to constrain the output of each ReLU within the section. The constraints used are $y>0$, $y>x$ and $y\le (u\cdot(x-l)) / (u-l)$ where $u$ and $l$ are the upper and lower bounds inferred from the boundaries of the input domain by considering the maximum and minimum values of each input variable.


\subsection{Building the Abstraction}

Section~\ref{sec:abstraction} describes our approach to defining an
abstract model of controller execution, as an MDP,
and Definition~\ref{def:ctrlabs} explains the structure required of this MDP
such that it can be solved to produce probabilistic guarantees,
i.e., upper bounds on the probability of a failure occurring within some time horizon $k$.
Here, we provide more details on the construction of the abstraction.

Algorithm~\ref{alg:buildmdp} shows pseudo code for the overall procedure.
We start from the initial abstract states $\hat{S}_0$, which are the initial
states of the MDP, and then repeatedly explore the ``frontier'' states,
whose transitions have yet to be constructed,
stopping exploration when either depth $k$ (the required time horizon) or
an abstract state containing a failure state is reached.
For each abstract state $\hat{s}$ to be explored,
we use the techniques from the previous section to split $\hat{s}$ into subregions
of states for which the controller policy selects the same action. 

Determining successor abstract states in the MDP uses the
environment abstraction $\hat{E}$ (see Definition~\ref{def:envabs}).
Since we use the ``Box'' abstract domain, this means using interval arithmetic,
i.e., computing the successors of the corner points enclosing the intervals
while the remaining points contained within them are guaranteed to be
contained within the enclosing successors.
The definitions of our concrete environments are therefore restricted to functions
that are extensible to interval arithmetic.

\begin{algorithm}[!t]
    \DontPrintSemicolon
    \SetKwFunction{FMain}{build\_mdp}
    \SetKwFunction{partition}{find\_action\_subregions}
    \caption{Build MDP}
    \label{alg:buildmdp}
    \SetAlgoLined
    \SetKwProg{Fn}{function}{:}{}
    \Fn{\FMain{$net$, $\hat{S}_0$}}{
        
        $\hat{S}_\mathit{frontier} = \hat{S}_0$, $t = 0$\;
        \While{$t<k$}{
            \ForEach{$\hat{s}\in \hat{S}_\mathit{frontier}$}{
                \ForEach{$\act\in\Act$}{
                
                $\hat{S}_\act,\hat{S}_{\overline{\act}}$ = \partition($net$, $\act$, $\hat{s}$)\;
                    \ForEach{$\hat{s}_j\in \hat{S}_\act$ and $p_i{:}w_i$ in $f(\act)$}{
                        $\hat{s}'$ = $\hat{E}(\hat{s}_j,w_i)$\;
                        store $(\hat{s},p_i,\hat{s}')$ in MDP\;
                        add $\hat{s}'$ to $\hat{S}_\mathit{frontier}$ unless $\hat{s}'\cap\mathit{fail}\neq\emptyset$\;
                    }
                }
            }
            $t=t+1$
        }
    }
\end{algorithm}

\subsection{Refining the Abstraction}

Although the MDP constructed as described above yields upper bounds
on the finite-horizon probability of failure, we can improve the results
by \emph{refining} the abstraction, i.e., further splitting some of the abstract states.
The refinement step aims to improve the precision of states which are considered unsafe
(assuming some specified probability threshold $p_\mathit{safe}$),
by reducing the upper bound closer to the real probability of encountering a failure state. 

Regions of initial abstract states that are considered unsafe
are split into smaller subregions and we then recreate the branches of the MDP abstraction
from these new subregions in the same way as described in Algorithm~\ref{alg:buildmdp}.
This portion of the MDP is then resolved, to produce a more accurate prediction
of their upper bound probability of encountering a failure state,
potentially discovering new safe subregions in the initial abstract state.
The refinement process is executed until either there are no more unsafe regions in the initial state or the maximum size of the intervals are less than a specified precision $\epsilon$.

\subsection{Storing and Manipulating Abstract States}

Very often abstract states have a topological relationship with other abstract states encountered previously.
One abstract state could completely encapsulate or overlap with another, but simply comparing all the possible pairs of states would be infeasible.
For this reason we need a data structure capable of reducing the number of comparisons to just the directly neighbouring states. 
A tree-like structure is the most appropriate and significant progress has been made on tree structures capable of holding intervals. However, most of them do not scale well for $n$-dimensional intervals with $n>3$.

R-tree~\cite{guttman84} is a data-structure that is able to deal with $n$-dimensional intervals, used to handle GIS coordinates in the context of map loading where only a specific area needs to be loaded at a time.
This data structure allows us to perform ``window queries'' which involve searching for $n$-dimensional intervals that intersect with the interval we are querying in O($\log_n(m)$) time, where $m$ is the number of intervals stored.
R-tree organises intervals and coordinates in nested ``subdirectories'' so that only areas relevant to the queried area are considered when computing an answer.

Here, we use an improved version of R-tree called R*-tree~\cite{beckmann1990r}
which reduces the overlapping between subdirectories at the cost of higher computational cost of O($n\log(m)$). This modification reduces the number of iterations required during the queries effectively speeding up the calculation of the results.
When an abstract domain is queried for the actions the controller would choose, only the areas which were not previously visited get computed.


\section{Experimental Results}

We have implemented our MOSAIC algorithm,
described in Sections~\ref{sec:maths} and~\ref{sec:algs},
and evaluated it on deep reinforcement learning controllers trained
on two different benchmark environments from
OpenAI Gym \cite{brockman2016openai}, a pendulum and a cartpole,
modified to include controller faults.
For space reasons, we consider only ``sticky'' actions~\cite{machado2018revisiting}:
each action is erroneously executed twice with probability $p=0.2$.

\startpara{Implementation}
Our implementation uses a combination of Python and Java.
The neural network architecture is handled through the Pytorch library~\cite{pytorch},
interval arithmetic with {\tt pyinterval}~\cite{pyinterval}
and graph analysis with {\tt networkX}~\cite{networkx}.
Abstract domain operations are performed with 
{\tt Rtree}~\cite{rtree}, building on the library {\tt libspatialindex}~\cite{libspatialindex}.
Constructing and solving MDPs
is done using PRISM~\cite{KNP11}, through its Java API,
built into a Python wrapper using {\tt py4j}~\cite{py4j}.

\subsection{Benchmarks and Policy Learning}


\startpara{Pendulum}
The pendulum environment consists of a pole pivoting around a point at one of its ends. The controller can apply a rotational force to the left or to the right with the aim of balancing the pole in its upright position.
The pole is underactuated which means that the controller can only recover to its upright position when the pole is within a certain angle. For this reason, if the pole goes beyond a threshold from which it cannot recover, the episode terminates and the controller is given a large negative reward.
Each state is composed of 2 variables: the angular position and velocity of the pole.

\startpara{Cartpole}
The cartpole environment
features a pole being balanced on top of a cart that can either move left or right.
The cartpole can only move within fixed bounds and the pole on top of it cannot recover its upright state after its angle exceeds a given threshold.
In this problem the size of each state is 4 variables: the position of the cart on the x-axis, the speed of the cart, the angle of the pole and the angular velocity of the pole.

\begin{figure}[!t]
	\centering
	
	\includegraphics[width=1\linewidth]{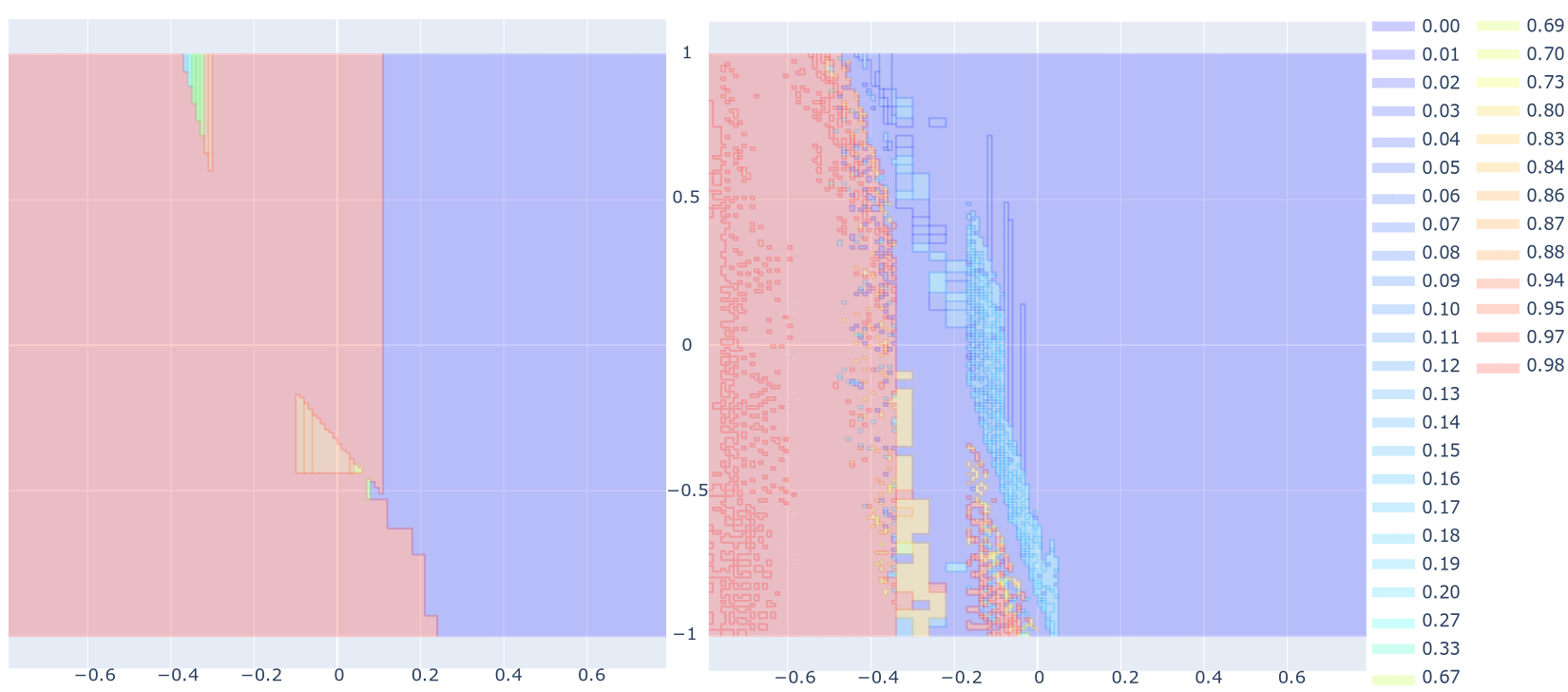}
    \caption{Heatmaps of failure probability upper bounds for subregions of initial states for the pendulum benchmark (x/y-axis: pole angle/angular velocity). Left: the initial abstraction; Right: the abstraction after 50 refinement steps.}
	\label{fig:pendulum_e2_h7_ub_merged}
	\vspace*{-0.2cm}
\end{figure}

\startpara{Policy construction}
We train our own controller policies for the benchmarks,
in order to take into account the controller failures added.
For the policy neural networks,
we use 3 fully connected layers of size 64, followed by an output layer
whose size equals the number of controller actions in the benchmark.
The training is performed by using the Deep Q-network algorithm \cite{Mnih2015} with prioritised experience replay~\cite{schaul2015prioritized}, which tries to predict the action value in each state and choosing the most valuable one.
For both environments, we train the controller for 6000 episodes, limiting the maximum number of timesteps for each episode to 1000. We linearly decay the epsilon in the first 20\% of the total episodes up to a minimum of 0.01 which we keep constant for the rest of the training. The remaining hyperparameters remain the same as suggested in \cite{Mnih2015} and \cite{schaul2015prioritized}.



\subsection{Results}
We have run the MOSAIC algorithm on the benchmark controller policies described above.
We build and solve the MDP abstraction to determine upper bounds on failure
probabilities for different parts of the state space.
Figure~\ref{fig:pendulum_e2_h7_ub_merged} (left) shows a heatmap of the probabilities for
various subregions of the initial states of the pendulum benchmark,
within a time horizon of 7 steps.
Figure~\ref{fig:pendulum_e2_h7_ub_merged} (right) shows the heatmap for a more precise abstraction,
obtained after 50 steps of refinement.
We do not fix a specific probability threshold $p_\mathit{safe}$ here,
but the right-hand part (in blue) has upper bound zero,
so is ``safe'' for any $p_\mathit{safe}>0$.
The refined abstraction discovers new areas which are safe
due to improved (i.e., lower) upper bounds in many regions.

\begin{figure}[!t]
\centering
\begin{minipage}{.5\linewidth}
\centering
\vspace*{10pt}
    \includegraphics[width=1\linewidth,height=160pt]{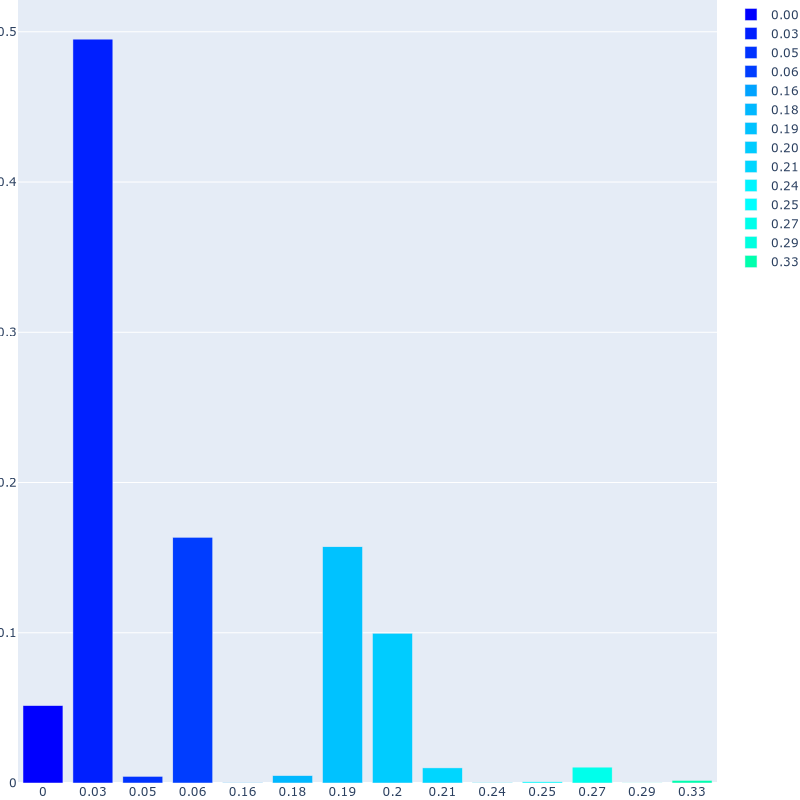}
	\caption{
		Cartpole: Histogram plot of the volume occupied by the initial state subregions, grouped by their maximum failure probability. \\ 
	}
	\label{fig:cartpole_e3_h5_p_chart}
\end{minipage}%
\begin{minipage}{.5\linewidth}
	\centering
    \includegraphics[width=1\linewidth,height=160pt]{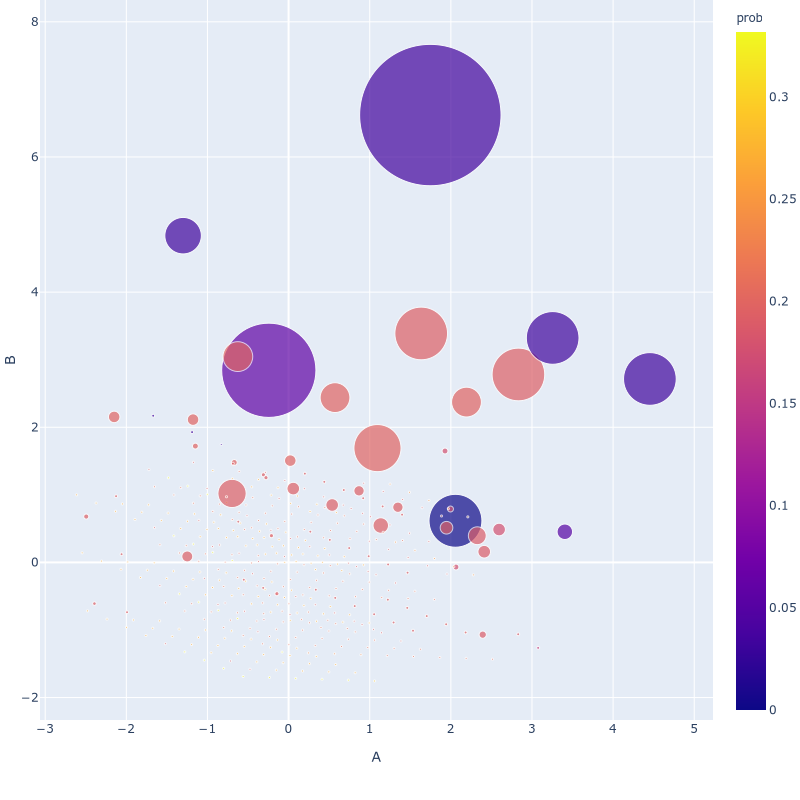}
    \vspace*{-1cm}
	\caption{
		Cartpole: probability bounds for initial state subregions
		(projection using principal component analysis;
		size denotes the volume occupied by the interval).
		We can see that large sections of the state space have max probability close to 0.
	}
	\label{fig:cartpole_e3_h7_pca}
\end{minipage}%

\end{figure}

Results for the cartpole example are harder to visualise
since the state space has 4 dimensions.
Figure~\ref{fig:cartpole_e3_h7_pca} shows a scatterplot of
failure probability bounds within 7 time steps
for the subregions of the initial state space;
the intervals have been projected to two dimensions using principal component analysis,
the size of the bubble representing the volume occupied by the interval.
We also plot, in Figure~\ref{fig:cartpole_e3_h5_p_chart},
a histogram showing how the probabilities are distributed
across the volume of the subregions of the initial states.
For a given value $p_\mathit{safe}$ on the x-axis,
our analysis yields a probabilistic guarantee of safety
for the sum of all volumes shown to the left of this point.

\startpara{Scalability and efficiency}
Lastly, we briefly discuss the scalabilty and efficiency of
our prototype implementation of MOSAIC.
Our experiments were run on a 4-core 4.2 GHz PC with 64 GB RAM running Ubuntu 18.04.
We successfully built and solved abstractions up to time horizons of 7 time-steps on both benchmark environments. For the pendulum problem, the size of the MDP built ranged up to approximately 160,000 states after building the initial abstraction, reaching approximately 225,000 states after 50 steps of refinement. For the cartpole problem, the number of states after 7 time-steps ranged up to approximately 75,000 states. The time required was roughly 50 minutes and 30 minutes for the two benchmarks, respectively.
%



\vspace*{-0.7em}
\section{Conclusions}

We have presented a novel approach called MOSAIC for verifying deep reinforcement learning systems
operating in environments where probabilistic controller faults may occur.
We formalised the verification problem as a finite-horizon analysis of a continuous-space
discrete-time Markov process and showed how to use a combination of abstract interpretation
and probabilistic model checking to compute upper bounds on failure probabilities.
We implemented our techniques and successfully applied them to two benchmark control problems.

Future work will include more sophisticated refinement and abstraction approaches,
including the use of lower bounds to better measure the precision of abstractions
and to guide their improvement using refinement.
We also aim to improve scalability to larger time horizons and more complex environments,
for example by investigating more efficient abstract domains.

\startpara{Acknowledgements}
This project has received funding from the European Research Council (ERC)
under the European Union’s Horizon 2020 research and innovation programme
(grant agreement No.~834115, FUN2MODEL).

\bibliographystyle{splncs03}
\bibliography{main}

\ifthenelse{\isundefined{\techreport}}{%
}{%

\newpage
\appendix
\section*{Appendix (Proof of Theorem~\ref{thm:abstr})}

We give here a proof of Theorem~\ref{thm:abstr}, from Section~\ref{sec:abstraction},
which states that:

\vskip1em

\noindent
Given a state $s\in S$ of a controller model DTMP,
and an abstract state $\hat{s}\in\hat{S}$ of the corresponding controller abstraction MDP
for which $s\in\hat{s}$, we have:
$$
\Pr{s}{}(\Diamond^{\leq k}\mathit{fail}) \ \leq \ \Pr{\hat{s}}{\max}(\Diamond^{\leq k}\mathit{fail})
$$
By the definition of $\Pr{\hat{s}}{\max}(\cdot)$, it suffices
show that there is \emph{some} policy $\sigma$ in the MDP such that:
\begin{equation}\label{eqn:sigma}
\Pr{s}{}(\Diamond^{\leq k}\mathit{fail}) \ \leq \ \Pr{\hat{s}}{\sigma}(\Diamond^{\leq k}\mathit{fail})
\end{equation}
Recall that, in the construction of the MDP (see Definition~\ref{def:ctrlabs}),
an abstract state $\hat{s}$ is associated with a partition of subsets $\hat{s}_j$
of $\hat{s}$, each of which is used to define the $j$-labelled choice in state $\hat{s}$.
Let $\sigma$ be the policy that picks in each state $s$ (regardless of history)
the unique index $j_s$ such that $s\in \hat{s}_{j_s}$.

The probabilities $\Pr{\hat{s}}{\sigma}(\Diamond^{\leq k}\mathit{fail})$
for this policy, starting in abstract state $\hat{s}$,
are defined similarly to those for discrete-time Markov processes
(see Section~\ref{sec:verif}):
$$
\Pr{\hat{s}}{\sigma}(\Diamond^{\leq k}\mathit{fail})=
\left\{\begin{array}{cl}
1 & \mbox{if } \hat{s}\models\mathit{fail} \\
0 & \mbox{if } \hat{s}\not\models\mathit{fail} \land k{=}0 \\
\sum\limits_{\hat{s}'\in \supp(\hat{\matr{P}}(\hat{s},j_s,\cdot))}\hat{\matr{P}}(\hat{s},j_s,\hat{s}'){\cdot}\Pr{\hat{s}'}{}(\Diamond^{\leq k-1}\mathit{fail}) & \mbox{otherwise.}
\end{array}\right.
$$
Since this is defined recursively, we prove (\ref{eqn:sigma}) by induction over $k$.

\vskip0.5em
\noindent
First, for the case $k=0$, from the computation of $\Pr{s}{}(\Diamond^{\leq k}\mathit{fail})$
(see Section~\ref{sec:verif}),
we have that $\Pr{s}{}(\Diamond^{\leq 0}\mathit{fail})$ is equal to 1 if $s\models\mathit{fail}$ and 0 otherwise.
The same holds for $\Pr{\hat{s}}{}(\Diamond^{\leq 0}\mathit{fail})$, as stated above.
From Definition~\ref{def:ctrlabs},
$s\models\fail$ implies $\hat{s}\models\fail$.
Therefore, $\Pr{s}{}(\Diamond^{\leq 0}\mathit{fail}) \ \leq \ \Pr{\hat{s}}{\sigma}(\Diamond^{\leq 0}\mathit{fail})$.

\vskip0.5em
\noindent
Next, for the inductive step, we will assume, as the inductive hypothesis,
that $\Pr{s'}{}(\Diamond^{\leq k-1}\mathit{fail}) \ \leq \ \Pr{\hat{s}'}{\sigma}(\Diamond^{\leq k-1}\mathit{fail})$ for $s'\in S$ and $\hat{s}'\in\hat{S}$ with $s'\in \hat{s}'$.
If $\hat{s}\models\fail$ then $\Pr{\hat{s}}{\sigma}(\Diamond^{\leq k}\mathit{fail})=1 \ \geq \ \Pr{s}{}(\Diamond^{\leq k}\mathit{fail})$.
Otherwise we have:
$$
\begin{array}{rcll}
&& \Pr{\hat{s}}{\sigma}(\Diamond^{\leq k}\mathit{fail}) \\

& = & \sum_{\hat{s}'\in \supp(\hat{\matr{P}}(\hat{s},j_s,\cdot))}\hat{\matr{P}}(\hat{s},j_s,\hat{s}')\cdot\Pr{\hat{s}'}{}(\Diamond^{\leq k-1}\mathit{fail}) \ \ & \mbox{by defn. of $\Pr{\hat{s}}{\sigma}(\Diamond^{\leq k}\mathit{fail})$} \\

& = & \sum_{w\in\Act^*}  f(\pi(\hat{s}_j))(w)\cdot\Pr{\hat{E}(\hat{s}_j,w)}{}(\Diamond^{\leq k-1}\mathit{fail}) & \mbox{by defn. of $\hat{\matr{P}}(\hat{s},j,\hat{s}')$} \\

& = & \sum_{w\in\Act^*}  f(\pi(s))(w)\cdot\Pr{\hat{E}(\hat{s}_j,w)}{}(\Diamond^{\leq k-1}\mathit{fail}) & \mbox{since $s\in \hat{s}_j$} \\

& \geq & \sum_{w\in\Act^*}  f(\pi(s))(w)\cdot\Pr{E(s,w)}{}(\Diamond^{\leq k-1}\mathit{fail}) & \mbox{by induction and since, by} \\
&&& \mbox{Defn.~\ref{def:envabs}, $E(s,w)\in\hat{E}(\hat{s}_j,w)$} \\

& = & \sum_{s'\in \supp(\matr{P}(s,\cdot))}\matr{P}(s,s')\cdot\Pr{s'}{}(\Diamond^{\leq k-1}\mathit{fail}) & \mbox{by defn. of $\matr{P}(s,s')$} \\[0.5em]

& = & \Pr{s}{}(\Diamond^{\leq k}\mathit{fail}) & \mbox{by defn. of $\Pr{s}{}(\Diamond^{\leq k}\mathit{fail})$} \\
\end{array}
$$
which completes the proof.

}%

\end{document}